\documentclass[preprint]{IEEEtran}

\usepackage[]{algorithm2e}
\usepackage{amsmath}
\usepackage{caption}
\usepackage{graphicx}
\usepackage[hidelinks]{hyperref}
\usepackage{makecell}
\usepackage{pgfplots,pgfplotstable}
\usepackage{subcaption}
\usepackage{tabularx}
\usepackage[normalem]{ulem}
\usepackage{xcolor}

\usepgfplotslibrary{dateplot}
\usepgfplotslibrary{external}
\usepgfplotslibrary{colorbrewer}
\pgfplotsset{
	compat=1.16,
	enlarge x limits=false,
	enlarge y limits=auto,
	cycle list/Paired,
	cycle list/Set1-5,
}
\pgfplotsset{every axis plot/.append style={line width=0.8pt}}

\definecolor{color1}{HTML}{003f5c}
\definecolor{color2}{HTML}{58508d}
\definecolor{color3}{HTML}{bc5090}
\definecolor{color4}{HTML}{ff6361}
\definecolor{color5}{HTML}{ffa600}
\definecolor{lightgray}{HTML}{fbfbfb}

\newcommand{\twodots}{\mathinner {\ldotp \ldotp}}

\begin{document}
	
	\title{Data-Driven Copy-Paste Imputation for Energy Time Series}
	
	\author{Moritz Weber, 
			Marian Turowski, 
			Hüseyin K. \c{C}akmak, 
			Ralf Mikut, 
			Uwe Kühnapfel, 
			Veit Hagenmeyer,~\IEEEmembership{Member,~IEEE}%
	\thanks{The present contribution is supported by the Helmholtz Association under the Joint Initiative ``Energy System 2050 - A Contribution of the Research Field Energy''}
	\thanks{M. Weber, M. Turowski, H. K. \c{C}akmak, R. Mikut, U. Kühnapfel, and V. Hagenmeyer are with the Institute for Automation and
	Applied Informatics, Karlsruhe Institute of Technology, 76344 Eggenstein-Leopoldshafen,
	Germany (e-mail: moritz.weber@kit.edu).}
	\thanks{The first two authors contributed equally to this work.}}

	\maketitle
	
	\begin{abstract}
		A cornerstone of the worldwide transition to smart grids are smart meters.
		Smart meters typically collect and provide energy time series that are vital for various applications, such as grid simulations, fault-detection, load forecasting, load analysis, and load management.
		Unfortunately, these time series are often characterized by missing values that must be handled before the data can be used.
		A common approach to handle missing values in time series is imputation.
		However, existing imputation methods are designed for \textit{power} time series and do not take into account the total energy of gaps, resulting in jumps or constant shifts when imputing \textit{energy} time series.
		In order to overcome these issues, the present paper introduces the new \textit{Copy-Paste Imputation (CPI)} method for \textit{energy} time series.
		The \textit{CPI} method copies data blocks with similar properties and pastes them into gaps of the time series while preserving the total energy of each gap.
		The new method is evaluated on a real-world dataset that contains six shares of artificially inserted missing values between 1 and 30\%.
		It outperforms by far the three benchmark imputation methods selected for comparison.
		The comparison furthermore shows that the \textit{CPI} method uses matching patterns and preserves the total energy of each gap while requiring only a moderate run-time.
	\end{abstract}


\section{Introduction and State of the Art}
\label{sec:intro}

In the course of the worldwide transition to an energy system mainly based on renewable energy sources, a key is the implementation of smart grids  \cite{Alahakoon2016SmartElectricity}.
Smart meters are a cornerstone of these smart grids and are thus installed in an increasing number worldwide.
They record and transmit a variety of data such as voltage, reactive power, or the electricity consumption of consumers \cite{Alquthami2020AnalyticsFramework}.
The collected data are an essential input to various applications supporting and enabling the transition to energy systems from renewable energies.
For example, the collected data allow grid operators to perform grid simulations \cite{Hagenmeyer2016InformationCommunication} for stability analysis, grid development, fault-detection, and efficiency improvements. 
The collected data is also needed for load forecasting \cite{Heidrich2020ForecastingEnergy}, load analysis, and load management \cite{Wang2019ReviewSmart}. 
Moreover, the collected data allow research facilities to develop technologies for the grid of the future.

The results of these applications highly depend on the quality of the input data.
The data quality, in turn, is highly influenced by two key challenges in the smart grid infrastructure: the \textit{accuracy} of data acquisition and the \textit{reliability} of transmission and storage \cite{King2012Chapter11}.
The accuracy of data acquisition refers to the correctness of the recorded data. It is reduced by problems causing, for example, noise and outliers in the data \cite{Chen2017DataQuality,Wang2020PointContextual}.
For further processing, outliers in particular are often detected and labeled as missing values as a first step \cite{Alquthami2020AnalyticsFramework, Akouemo2017DataImproving}.
The reliability of transmission and storage, however, mainly relates to the completeness of the recorded data.
In implemented smart meter systems, recorded data contain between 3 and 4\% of missing values, for example due to planned outages \cite{King2012Chapter11,Peppanen2015LeveragingAMI}.
Due to these two key challenges of smart grids, missing values in recorded data are a common problem.
Although some applications are able to handle incomplete data \cite{Taylor2018ForecastingScale}, most applications require that the missing values are handled by pre-processing the data.

A common method to handle missing data is \textit{imputation}.
Imputation replaces missing values with values that should resemble the actual data \cite{Moritz2017ImputeTSTime}.
Since missing values are a common problem in real-world datasets, many imputation methods exist in the literature for time series in general:
Imputation methods range from very basic methods such as linear interpolation and \textit{Last Observation Carried Forward (LOCF)} \cite{Moritz2017ImputeTSTime} over time series analysis-based methods  \cite{Akouemo2017DataImproving,Akouemo2014TimeSeries} to learning-based methods  \cite{Cao2018BRITSBidirectional,Bokde2018NovelImputation}.

Since the imputation of general time series can be a very difficult problem -- especially without additional information --, focusing on a certain type of time series allows to enhance the imputation.
The enhancement can be, for example, related to the characteristics of the considered type of time series.
In the context of smart meters, the time series of recorded electricity consumption or generation typically depend on factors such as weather, human routines, social norms (e.g. weekends or holidays) and many others \cite{Peppanen2016HandlingBad, Ordiano2018EnergyForecasting}.
These factors often lead to the commonly known patterns with different periodicity -- mostly daily, weekly, and yearly.

Utilizing these properties of smart meter time series is a common approach because it enables an imputation without additional data. 
For example, \cite{Friese2013UniFIeDUnivariate} utilize the existence of daily and weekly patterns.
In their work, they estimate the pattern frequency of the time series using the auto-correlation function and use the mean values of the estimated pattern frequency to impute missing values.
In another work \cite{Matheson2004MeterData}, the similarity between days is used by filling larger gaps with the average values of validated reference days.
Very short gaps with a length of 2 hours or less are imputed with linear interpolation, as this often fits the very short-term characteristics of smart meter time series.
The \textit{Optimally Weighted Average} approach in \cite{Peppanen2016HandlingBad}, utilizes daily and weekly patterns as well as seasonality to select appropriate historical values.
These values are used to calculate historical averages that are combined with linear interpolation for smooth transitions between actual and imputed values.
In \cite{Mateos2013LoadCurve}, a method for imputation, de-noising, and outlier removal based on \textit{Principal Component Pursuit} is introduced that utilizes the spatial correlations of load profiles of adjacent substations.
Similarly, in \cite{Borges2020EnhancingMissing} the imputation of substation data is formulated as a forecasting problem, utilizing collected data of nearby substations as well as weather data, which often has an impact on power consumption and generation.

While all of these imputation approaches are specifically designed for smart meter time series, they are limited to the imputation of \textit{power} time series.
In a \textit{power} time series, every entry contains the average power consumption or generation between two successive time steps.
However, smart meters typically provide \textit{energy} time series by default. In an \textit{energy} time series, every entry contains the meter reading, i.e.~the energy that has been consumed or generated up to this point of time.
Therefore, -- unlike in power time series -- if entries between two time steps are missing in an \textit{energy} time series, the next existing entry still contains the information about the total energy, which was consumed or produced during the missing values.
As a consequence, a \textit{power} time series can be derived from an \textit{energy} time series with missing values but not vice versa.

Thus, in the present paper, we propose the novel \textit{Copy-Paste Imputation (CPI)} method for univariate \textit{energy} time series.
It copies blocks of data with similar properties into gaps.
By copying blocks of matching data, the inherent patterns of the time series are preserved.
Even in time series with irregular pattern changes, considering the information about the total energy of each gap allows a matching selection of the blocks.
The method utilizes the information that energy time series contain as opposed to power time series and can, therefore, guarantee that the total recorded energy remains unchanged.
To the best of our knowledge, no other method in literature has so far used energy time series for imputation.

The remainder of the present paper is structured as follows.
The proposed method is explained in detail in \autoref{sec:cpi} and evaluated against three benchmark methods on a real-world dataset in \autoref{sec:evaluation}.
Concluding remarks and an outlook are given in \autoref{sec:conclusion}.

\section{Novel Copy-Paste Imputation Method}
\label{sec:cpi}

\begin{figure*}
	\colorbox{lightgray}{
	\begin{subfigure}{\linewidth}
		\begin{minipage}[c]{0.01\linewidth}
			(a)
		\end{minipage}
		\hfill
		\begin{minipage}[c]{0.63\linewidth}
			\includegraphics[width=\linewidth,trim={.6cm 0 0 0},clip]{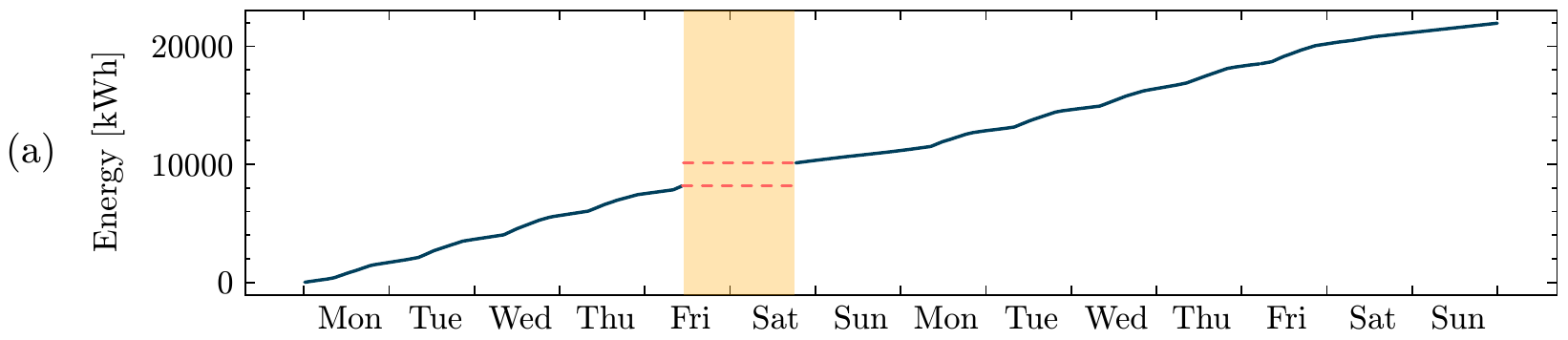}
		\end{minipage}
		\hfill
		\begin{minipage}[c]{0.3\linewidth}
			\caption{
				Initial energy consumption time series with a gap, i.e.~missing values on the first Friday and Saturday, in yellow. The dashed lines indicate the energy difference between the last known value before and first known value after the gap.
			}
			\label{subfig:cpi-a}
		\end{minipage}
	\hfill
	\end{subfigure}}
	\vskip 12pt
	\colorbox{lightgray}{
	\begin{subfigure}{\linewidth}
		\begin{minipage}[c]{0.01\linewidth}
			(b)
		\end{minipage}
		\hfill
		\begin{minipage}[c]{0.63\linewidth}
			\includegraphics[width=\linewidth,trim={.6cm 0 0 0},clip]{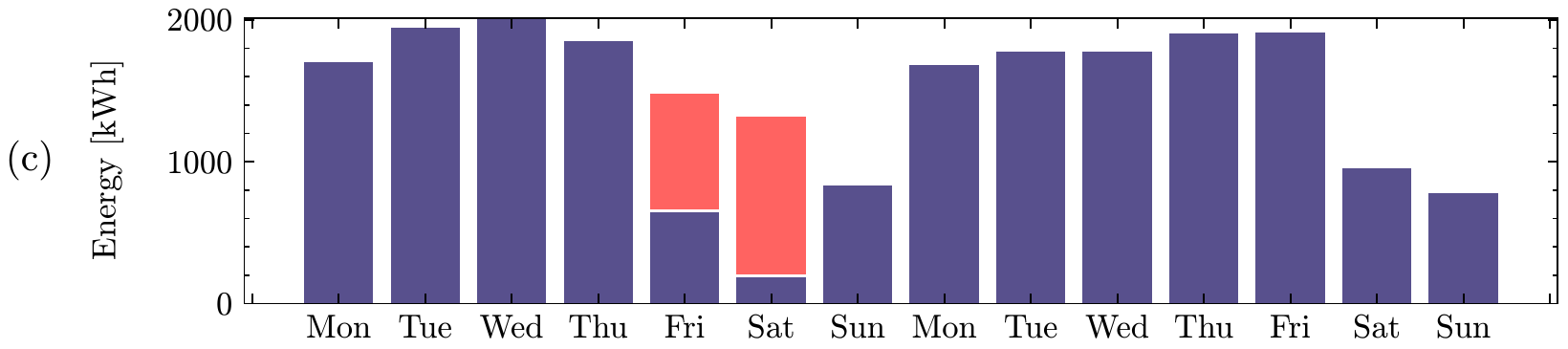}
		\end{minipage}
		\hfill
		\begin{minipage}[c]{0.3\linewidth}
			\caption{
			Energy consumption per day with an estimated consumption in red for the days with missing values proportional to their share of missing values.
			}
		\label{subfig:cpi-b}
		\end{minipage}
	\hfill
	\end{subfigure}}
	\colorbox{lightgray}{
	\begin{subfigure}{\linewidth}
		\begin{minipage}[c]{0.01\linewidth}
			(c)
		\end{minipage}
		\hfill
		\begin{minipage}[c]{0.63\linewidth}
			\includegraphics[width=\linewidth,trim={.6cm 0 0 0},clip]{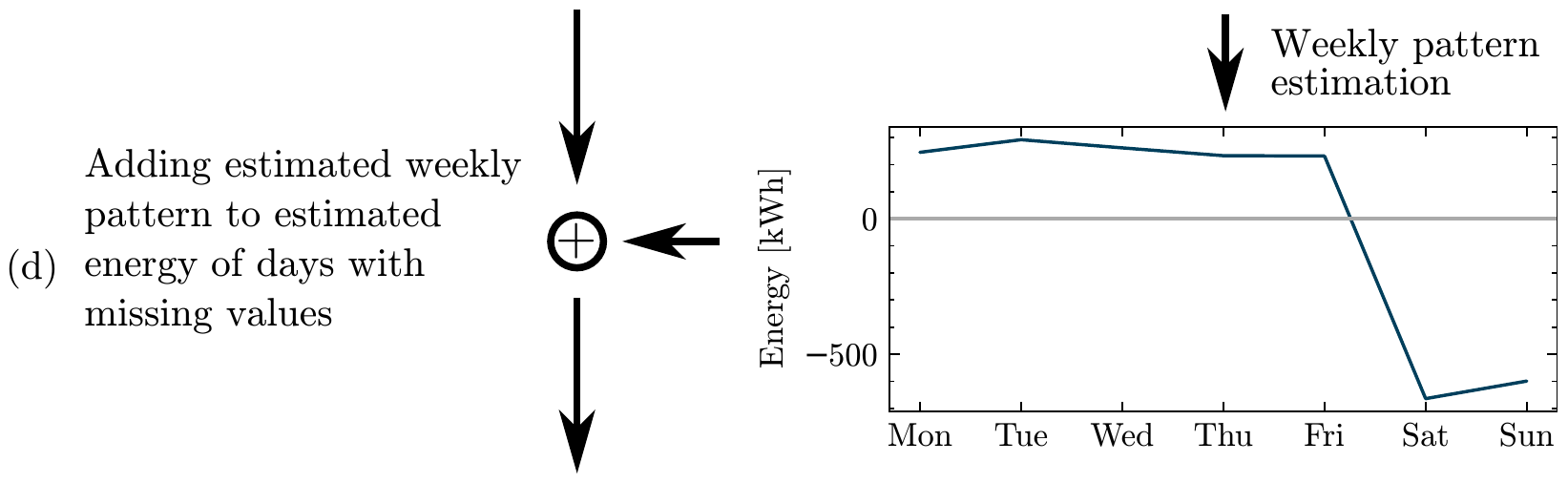}
		\end{minipage}
		\hfill
		\begin{minipage}[c]{0.3\linewidth}
			\caption{
				Estimated weekly pattern of the daily energy consumption that is added to the estimated energy consumption per day in (c), resulting in a new estimated energy consumption per day in (e).
			}
		\label{subfig:cpi-c}
		\end{minipage}
	\hfill
	\end{subfigure}}
	\colorbox{lightgray}{
	\begin{subfigure}{\linewidth}
		\begin{minipage}[c]{0.01\linewidth}
			(d)
		\end{minipage}
		\hfill
		\begin{minipage}[c]{0.63\linewidth}
			\includegraphics[width=\linewidth,trim={.6cm 0 0 0},clip]{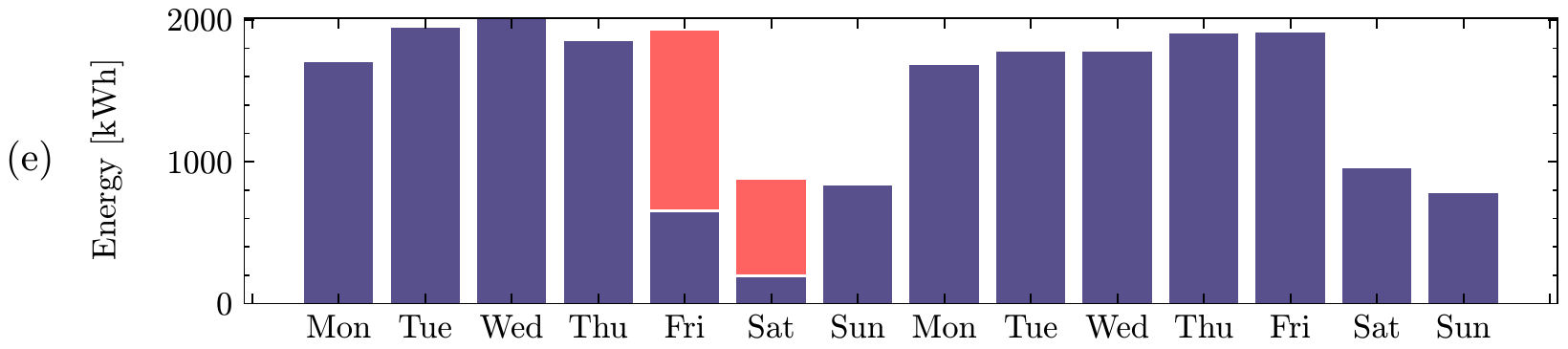}
		\end{minipage}
		\hfill
		\begin{minipage}[c]{0.3\linewidth}
			\caption{
				Energy consumption per day with an estimated consumption according to the weekly pattern in red for the days with missing data.
			}
			\label{subfig:cpi-d}
		\end{minipage}
		\hfill
	\end{subfigure}}
	\vskip 12pt
	\colorbox{lightgray}{
	\begin{subfigure}{\linewidth}
		\begin{minipage}[c]{0.01\linewidth}
			(e)
		\end{minipage}
		\hfill
		\begin{minipage}[c]{0.63\linewidth}
			\includegraphics[width=\linewidth,trim={.6cm 0 0 0},clip]{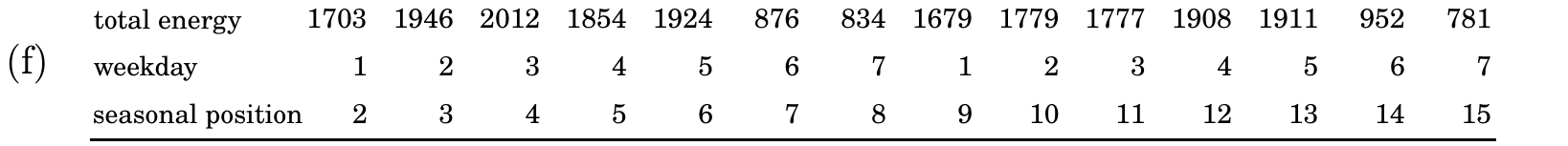}
		\end{minipage}
		\hfill
		\begin{minipage}[c]{0.3\linewidth}
			\caption{
				Properties of all shown days.
			}
		\label{subfig:cpi-e}
		\end{minipage}
		\hfill
	\end{subfigure}}
	\colorbox{lightgray}{
	\begin{subfigure}{\linewidth}
		\begin{minipage}[c]{0.01\linewidth}
			(f)
		\end{minipage}
		\hfill
		\begin{minipage}[c]{0.63\linewidth}
			\includegraphics[width=\linewidth,trim={.6cm 0 0 0},clip]{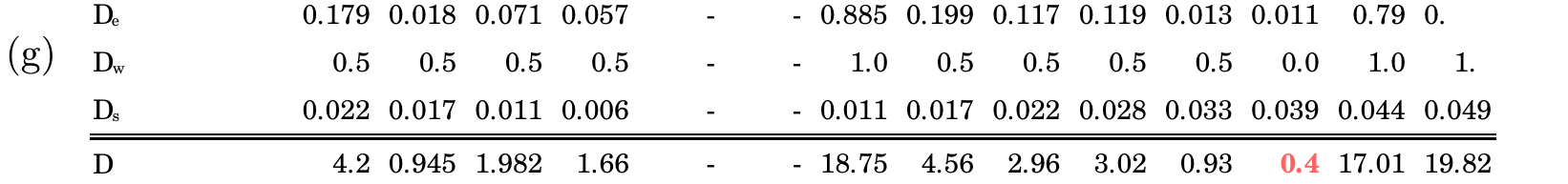}
		\end{minipage}
		\hfill
		\begin{minipage}[c]{0.3\linewidth}
			\caption{
				Resulting distances between the first day with missing values $d_4$, i.e.~the first Friday, and the other days.
				The distance calculation is parameterized with $w = (20, 1, 5)$.
			}
		\label{subfig:cpi-f}
		\end{minipage}
		\hfill	
	\end{subfigure}}
	\vskip 12pt
	\colorbox{lightgray}{
	\begin{subfigure}{\linewidth}
		\begin{minipage}[c]{0.01\linewidth}
			(g)
		\end{minipage}
		\hfill
		\begin{minipage}[c]{0.63\linewidth}
			\includegraphics[width=\linewidth,trim={.6cm 0 0 0},clip]{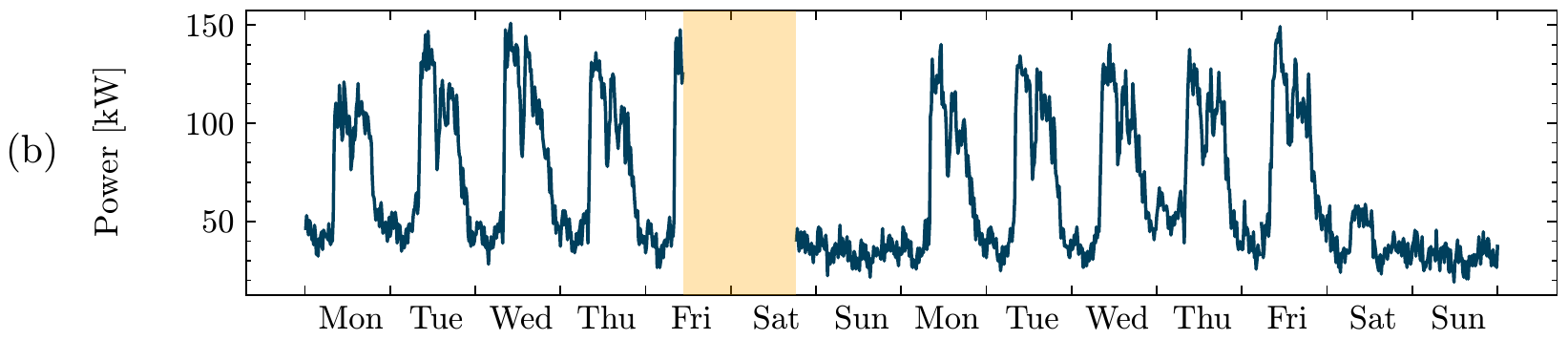}
		\end{minipage}
		\hfill
		\begin{minipage}[c]{0.3\linewidth}
			\caption{
				The power time series derived from (a) showing typical patterns of a load curve.
			}
		\label{subfig:cpi-g}
		\end{minipage}
		\hfill
	\end{subfigure}}
	\colorbox{lightgray}{
	\begin{subfigure}{\linewidth}
		\begin{minipage}[c]{0.01\linewidth}
			(h)
		\end{minipage}
		\hfill
		\begin{minipage}[c]{0.63\linewidth}
			\includegraphics[width=\linewidth,trim={.6cm 0 0 0},clip]{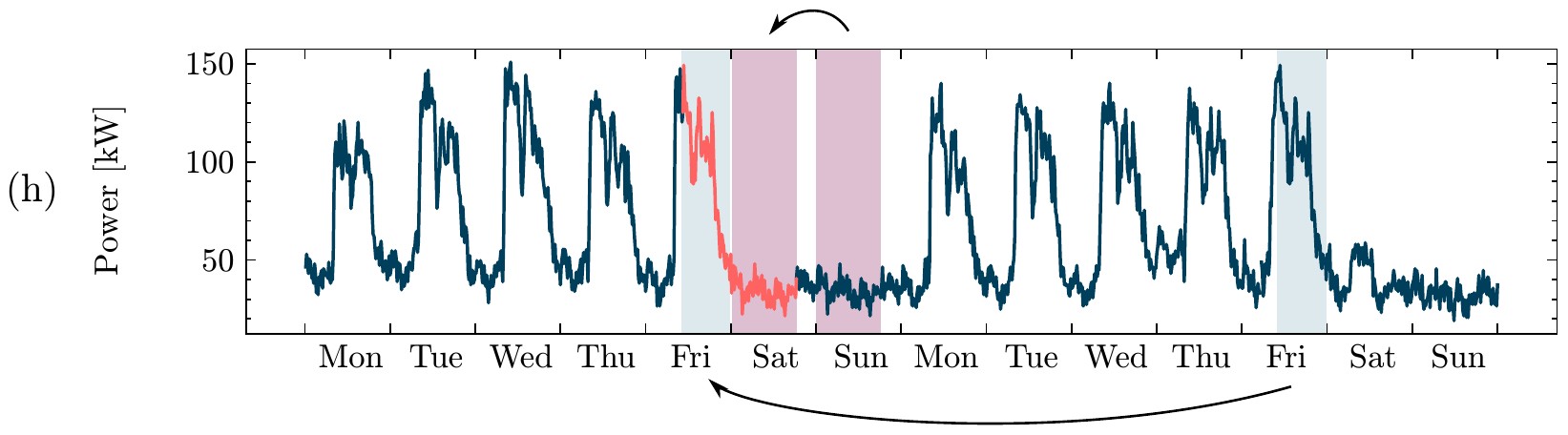}
		\end{minipage}
		\hfill
		\begin{minipage}[c]{0.3\linewidth}
			\caption{
				Summary of the procedure of the CPI method that imputes the missing values on Friday and Saturday.
				As indicated in (f), the second Friday with $D = 0.4$ is selected to fill the first gap.
			}
		\label{subfig:cpi-h}
		\end{minipage}
		\hfill
	\end{subfigure}}
	\caption{
		Illustration of the novel Copy-Paste Imputation method for two weeks of a real-world energy consumption time series with missing values.
	}
	\label{fig:cpi-example}
\end{figure*}

In this section, the newly proposed \textit{Copy-Paste Imputation (CPI)} method\footnote{A \textit{Python} implementation of the \textit{CPI} method is available on \url{https://github.com/KIT-IAI/CopyPasteImputation}.} is described. 
As illustrated in \autoref{fig:cpi-example}, the \textit{CPI} method uses an energy time series with missing values as input and imputes the missing values by filling them with the best matching days of the same time series.

\subsection{Linear Interpolation of Single Missing Values}
\label{subsec:linear-interpolation}

In the first step of the \textit{CPI} method, single missing values are imputed. A linear interpolation is used for this purpose because it provides sufficiently correct estimates for individual missing values.
The imputed values are considered as correct in the subsequent steps in order to increase the number of days without missing values that are available for copying.

\subsection{Energy Consumption Estimation}
\label{subsec:consumption-estimation}

The second step of the \textit{CPI} method is the energy consumption estimation for days with gaps.
The total energy consumption%
\footnote{In the following, we refer to consumption data only, but the same principles apply to generation data.}
during gaps is calculated by subtracting the last known energy value before the gap from the first known energy value after the gap (see \autoref{subfig:cpi-a}).

However, for gaps longer than one day, the calculated energy consumption must be allocated to the respective days appropriately.
For this purpose, firstly, the calculated energy consumption of the gap is distributed to the respective days according to their share of missing values.
\autoref{subfig:cpi-b} illustrates this distribution of the calculated energy consumption between Friday and Saturday for the given example.
Secondly, the \textit{Prophet} method \cite{Taylor2018ForecastingScale} for time series forecasting is used to estimate a weekly pattern in the daily energy consumption, utilizing only the days without missing values.
For the given example, \autoref{subfig:cpi-c} visualizes such a weekly pattern.
Thirdly, the estimated weekly pattern is added to the previously allocated energy consumption without changing the total energy consumption of each gap as shown in \autoref{subfig:cpi-d}.

\subsection{Compilation of Available Complete Days}

In the third step of the \textit{CPI} method, a list of the available complete days (i.e. days without missing values) is compiled. 
Assuming daily patterns, a weekly cycle, and a yearly seasonality in the energy consumption, each day is listed with its following properties: its total energy consumption ($d_e$), its weekday ($d_w \in \{1 \twodots 7\}$), and its seasonal position ($d_s$).
Under the assumption of a yearly seasonality, i.e.~365 days or 366 days for leap years, it follows that $d_s$ is in $\{1 \twodots 366\}$.
An example of a list with days and their properties is shown in \autoref{subfig:cpi-e}.

\subsection{Calculation of Dissimilarity Between Days}
\label{subsec:dissimilarity-calculation}

In the fourth step, the \textit{CPI} method calculates the dissimilarity to all complete days for each day with gaps using the days' previously introduced properties total energy, weekday, and seasonal position. 
For this reason, three distance measures, i.e.~$D_e$, $D_w$, and $D_s$, are calculated for each day with gaps $d_i$ and each available complete day $d_j$ after determining the properties of the respective day with missing values.

The first distance measure $D_e$ describes the distance between the total energy consumption of a day with gaps $d_i$ and a complete day $d_j$.
It is defined as
\begin{equation}\label{eq:dist-consumption}
D_e(d_i, d_j) = \frac{|d_{i,e} - d_{j,e}|}{e_{max} - e_{min}},
\end{equation}
where $e_{max}$ and $e_{min}$ are the maximum and minimum energy consumption of a day in the time series and $d_{i,e}$ and $d_{j,e}$ are the total energy consumption of the days $d_i$ and $d_j$.
For $d_i$, i.e.~the day with gaps, the previously estimated energy consumption is used.
Dividing by the difference between $e_{max}$ and $e_{min}$ ensures that the distance measure $D_e$ is in $[0, 1]$.

The second distance measure $D_w$ is based on the assumption of a weekly pattern in the time series and describes the distance between the weekday of a day with gaps $d_i$ and a complete day $d_j$. 
It is defined as
\begin{equation}\label{eq:dist-weekday}
D_{w}(d_i, d_j) = \left\{\begin{array}{ll}

0.0, & \text{if } d_{i,w} = d_{j,w}\\

0.5, & \text{if } d_{i,w} \in \{1..5\} \wedge d_{j,w} \in \{1..5\}\\
&
\vee d_{i,w} \in \{6,7\} \wedge d_{j,w} \in \{6,7\}\\
1.0, & \text{else},
\end{array}\right.
\end{equation}
where $d_{i, w}$ and $d_{j, w}$ are integer representations for the weekday of days $d_i$ and $d_j$.
One to five represent the workdays Monday to Friday, whereas 6 and 7 represent the weekend days Saturday and Sunday.
This distance measure $D_w$ assigns smaller distances to days of the same weekday or days of the same class (i.e.~workday or weekend) and higher distances to days of different classes.

The third distance measure $D_s$ captures the underlying seasonal patterns and describes the distance between the seasonal position of a day with gaps $d_i$ and a complete day $d_j$. It is defined as
\begin{equation}\label{eq:dist-year}
D_{s}(d_i, d_j) = \left\{\begin{array}{ll}

\frac{|d_{i,s} - d_{j,s}|}{\lfloor\frac{\mathbf{s}}{2}\rfloor}, & \text{if } |d_{i,y} - d_{j,y}| \le \lfloor\frac{\mathbf{s}}{2}\rfloor\\[2mm]
\frac{\mathbf{s} - |d_{i,s} - d_{j,s}|}{\lfloor\frac{\mathbf{s}}{2}\rfloor}, & \text{else},
\end{array}\right. 
\end{equation}
where $\mathbf{s}$ is the length of the seasonal cycle and $d_{i,s}$ and $d_{j,s}$ are the position of days $d_i$ and $d_j$ in this cycle.
For a yearly seasonality, $\mathbf{s}$ can be set to 365 or 366 to reflect the number of days in a year.
This distance measure ensures that two days from the same season are considered as more similar than two days from different seasons.
For example, January 1 and December 31 of the same year are almost one year apart but have a minimal distance $D_s$.
In contrast, January 1 and July 1 are only half a year apart and have a maximal distance $D_s$.

In order to determine the dissimilarity between a day with gaps and a complete day, the three individual distance measures are combined into a single criterion.
The resulting dissimilarity criterion $D$ is the weighted sum of the three individual distance measures $D_e$, $D_w$, and $D_s$. It is defined as
\begin{equation}\label{eq:dist}
D = w_e D_e + w_w D_w + w_s D_s,
\end{equation}
where $w_e$, $w_w$, and $w_s$ are the weights and $D_e$, $D_w$, and $D_s$ are the normalized distance measures. 
The individual distance measures are normalized to the interval [0,1] for an easier interpretation of these weights.
To determine the weights, for example, a grid search can be applied (see \hyperref[subsub:weights]{Part 3 of Subsection~\ref*{subsec:experimental-setting}} for an example).
For the given example, \autoref{subfig:cpi-f} shows the individual distances between the first Friday as a day with gaps and all the other days and the resulting dissimilarity values $D$ for given weights.

\subsection{Copy and Paste of Matching Days}
\label{subsec:imputing-gaps}

In the last step, the \textit{CPI} method copies the best matching days, pastes them into gaps, and scales the imputed values to preserve the energy of the respective gaps.
In order to determine the best matching days, the previously generated list of complete days is used.
For a day with gaps $d_i$, the day $d_j$ with the smallest dissimilarity $D(d_i , d_j )$ is chosen.
Since the entire list of complete days is used, days from the future of the day with gaps are also considered.
In the given example in \autoref{subfig:cpi-f}, the most similar day is the second Friday of the time series because that Friday has the lowest dissimilarity value.

Based on the determined best matching days, the actual copying and pasting of the best matching days into gaps is done.
For this purpose, the power time series, as shown in \autoref{subfig:cpi-g}, serves as basis. It can be derived from the input energy time series by calculating the average power $p_t$ between time steps $t-1$ and $t$, i.e.

\begin{equation}\label{eq:energy-to-power}
	p_t = \frac{e_t - e_{t-1}}{\Delta t},
\end{equation}

where $\Delta t$ is the time between two time steps, $e_t$ and $p_t$ are the energy and power at time step $t$, and $e_{t-1}$ is the energy at time step $t-1$.
In the derived power time series, every missing value in each day with gaps is replaced by the corresponding value of the previously determined best matching complete day (see \autoref{subfig:cpi-h}).

Finally, the imputed power values are scaled in order to preserve the actual energy of each gap.
For this purpose, the actual energy and the imputed energy are determined.
The actual energy $E_i$ of the gap $i$ from time step $t$ to time step $t+k$ is calculated as the energy difference, i.e.
\begin{equation}\label{eq:actual-energy}
	E_i = e_{t+k} - e_{t-1},
\end{equation}
where $e_{t+k}$ and $e_{t-1}$ are the energy at the time steps $t+k$ and $t-1$.
The imputed energy $E'_i$ is calculated by accumulating the imputed power values.
To preserve the energy, the imputed power values of gap $i$ are multiplied with the ratio of the actual energy and the imputed energy, i.e.
\begin{equation}\label{eq:scaled-imputation}
	\hat{p}_t = \hat{p}'_t \cdot \frac{E_i}{E'_i},
\end{equation}
where $\hat{p}'_t$ is the power value created by the copy-paste mechanism and $\hat{p}_t$ is the scaled power value.
The power time series completed in this way can then be used to calculate a complete energy time series by solving \autoref{eq:energy-to-power} for $e_t$.

\section{Evaluation}
\label{sec:evaluation}

In this section, the proposed \textit{CPI} method is evaluated on real-world data and its performance is compared to benchmark methods.
Therefore, the used dataset is introduced followed by the benchmark methods. 
After describing the experimental setting, the results are presented.

\subsection{Dataset}
\begin{figure*}
	\centering
	\includegraphics[width=\linewidth]{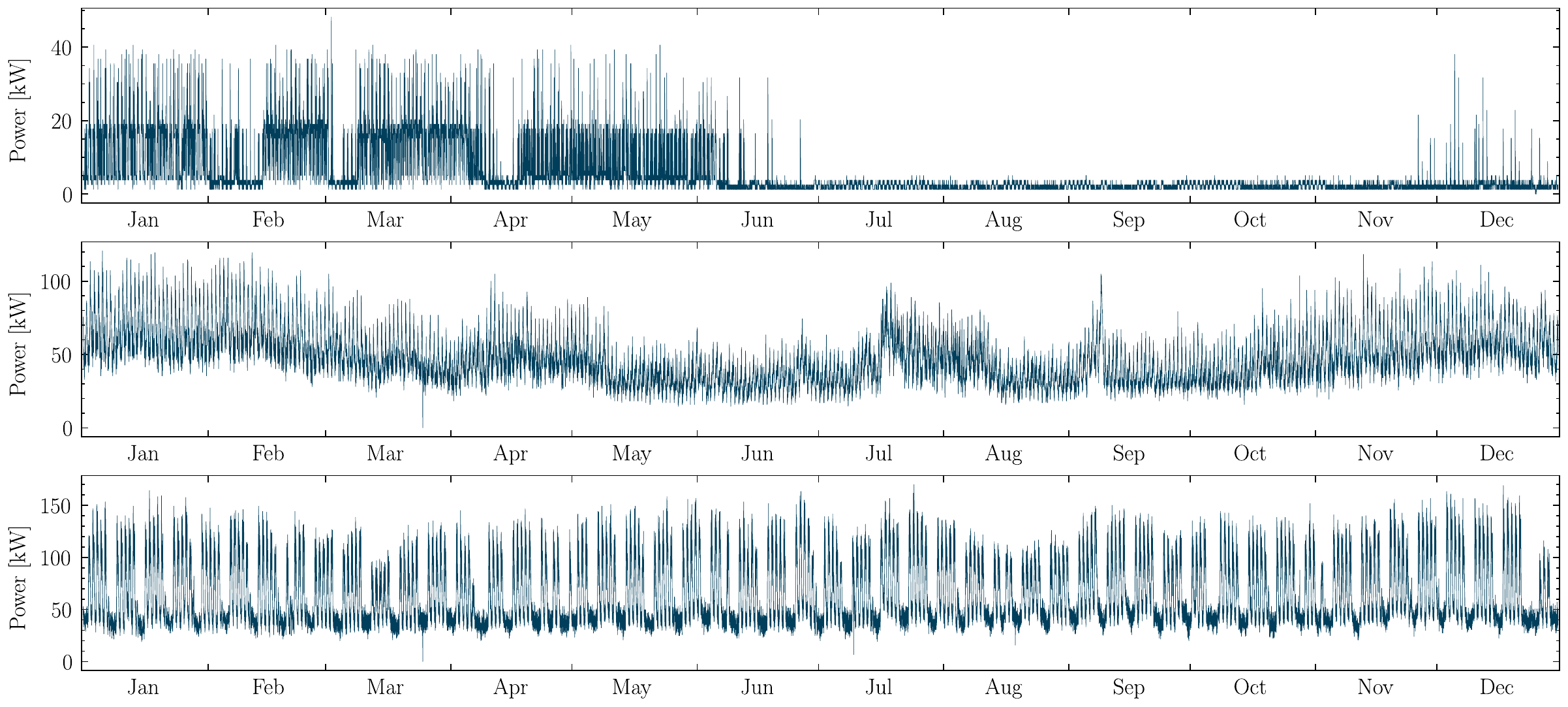}
	\caption{Three exemplary time series from the \textit{UCI} dataset, containing variations of different seasonal, weekly, and daily patterns.}
	\label{fig:uci-examples}
\end{figure*}

The dataset used for the evaluation is the \textit{ElectricityLoadDiagrams20112014} dataset%
\footnote{\url{https://archive.ics.uci.edu/ml/datasets/ElectricityLoadDiagrams20112014}}
from the \textit{UCI Machine Learning Repository} \cite{Dua2019UCIMachine}.
The dataset consists of power time series with complete consumption data from 370 different smart meters over a period of up to four years.
The time series contain quarter-hourly average power values in kW, resulting in 35,040 values per year.
Of these 370 time series, 50 differently shaped time series with a length of one year are selected as a representative sample.
The selected time series vary greatly in terms of seasonal, weekly, and daily patterns as illustrated in \autoref{fig:uci-examples}.

For the evaluation of the \textit{CPI} method, the selected power time series that do not contain any missing values are converted to energy time series by accumulating the power values.
Due to the completeness of the used time series, we insert artificially missing values by removing values from the time series.
The removal is parameterized with the maximum gap length and the share of missing values in the time series.
For the evaluation, six shares of missing values are used between 1\% and 30\%, i.e.~1, 2, 5, 10, 20, and 30\%.
In order to consider both larger gaps and single missing values, 5\% of each share of missing values are single missing values.

\subsection{Benchmark Methods}
In order to compare the performance of the proposed \textit{CPI} method, we apply benchmark methods to the dataset.
Due to the lack of imputation methods for energy time series -- to the best knowledge of the authors --, we consider imputation methods for power time series and time series in general.
Methods requiring additional data such as weather data \cite{Akouemo2017DataImproving,Akouemo2014TimeSeries} or validated reference days \cite{Matheson2004MeterData} and methods designed for multivariate time series only \cite{Cao2018BRITSBidirectional,Mateos2013LoadCurve,Borges2020EnhancingMissing} are discarded due to their lack of comparability.
Furthermore, during the evaluation, the method in \cite{Bokde2018NovelImputation} is excluded due to its excessive run-time.

In this context, we select three methods as benchmarks in view of comparison complexity and fairness.
We derive these methods from literature \cite{Taylor2018ForecastingScale,Moritz2017ImputeTSTime,Peppanen2016HandlingBad,Friese2013UniFIeDUnivariate} and adapt them where necessary.

The first benchmark method is a commonly applied linear interpolation \cite{Moritz2017ImputeTSTime,Peppanen2016HandlingBad}.
This method represents a lower baseline and should be outperformed in any case.
It imputes missing values $\hat{p}_t$ by linearly interpolating the first and last known power value before and after a gap, i.e.
\begin{equation}
	\hat{p}_t = \frac{t - t_1}{t_2 - t_1} \cdot (p_{t_2} - p_{t_1}) + p_{t_1},
\end{equation}
where $t_1$ and $t_2$ are the time steps before and after the gap.

The second benchmark method is the historical average (HistAvg) method similar to \cite{Friese2013UniFIeDUnivariate,Peppanen2016HandlingBad}.
Assuming a weekly pattern, this method calculates an average week based on all weeks available in the dataset.
It imputes missing values $\hat{p}_t$ with the corresponding values from that average week $\bar{p}_s$, i.e.
\begin{equation}
	\hat{p}_t = \bar{p}_s,
\end{equation}
where $s = t \mod W$ with $W$ being the number of power values per week.

The third benchmark method is based on the \textit{Prophet} model for time series forecasting \cite{Taylor2018ForecastingScale}.
\textit{Prophet} is a modular regression model that can be described as
\begin{equation}
	y(t) = g(t) + s(t) + h(t) + \varepsilon_t, 
\end{equation}
where $g$ is a model for the trend, $s$ for seasonality, $h$ for holidays, and $\varepsilon_t$ for changes that are not represented in the model.
The imputation method based on this model exploits \textit{Prophet}'s capability to estimate a time series model on irregularly spaced data \cite{Taylor2018ForecastingScale} and imputes missing values with the corresponding values of the model.

\subsection{Experimental Setting}
\label{subsec:experimental-setting}

This subsection describes the used hard- and software platform, introduces the error measures used to evaluate the imputation methods, and explains the weights in the dissimilarity measure of the evaluated \textit{CPI} method.

\subsubsection{Hard- and Software Platform}
For the evaluation of the \textit{CPI} and the benchmark methods, we compare the quality of the imputation and the required run-time.
For a better comparability of the results, all methods are implemented in \textit{Python} and evaluated on the same hardware.
The evaluation hardware is a notebook running \textit{Ubuntu 20.04} with an \textit{Intel Core i7-4770HQ} processor and 16GB of memory.

\subsubsection{Error Measures}
In order to evaluate the quality of an imputation in energy time series, we examine both the use of matching patterns to fill gaps and the conservation of the total energy in the gaps.
To evaluate the use of matching patterns, we determine how well imputed patterns match the actual patterns.
For this purpose, we measure the deviation between every single actual power value and the corresponding imputed power value using the \textit{Mean Absolute Percentage Error (MAPE)}. 
It is defined as 
\begin{equation}\label{eq:MAPE_p}
	\mathrm{MAPE_p} = \frac{1}{|T_m|} \Sigma_{t \in T_m} \left| \frac{\hat{p}_t - p_t}{p_t} \right| ,
\end{equation}
where $p_t$ and $\hat{p}_t$ are the actual and imputed power values at time step $t$ and $T_m$ is the set of time steps with missing values.
To evaluate the conservation of the total energy in gaps, we measure the difference between the actual and imputed energy while ignoring the fine granular patterns that are used for the imputation.
The difference is determined using the \textit{Weighted Absolute Percentage Error (WAPE)}, which is defined as
\begin{equation}\label{eq:WAPE_E}
	\mathrm{WAPE_E} = \frac{\Sigma_{i=1}^N |\hat{E}_i - E_i|}{\Sigma_{i=1}^N E_i},
\end{equation}
where $E_i$ and $\hat{E}_i$ are the actual and imputed energy of gap $i$ in a time series with $N$ gaps.
In contrast to the $\mathrm{MAPE_p}$ (\ref{eq:MAPE_p}), the weighting of the individual absolute errors is necessary in the $\mathrm{WAPE_E}$~(\ref{eq:WAPE_E}) to account for gaps of different sizes.

\subsubsection{Weights of the CPI Dissimilarity Measure}
\label{subsub:weights}
In order to apply the \textit{CPI} method, it is necessary to determine the weights for the three distance measures regarding total energy consumption, weekday, and seasonal position, which are used to calculate the dissimilarity between two days.
To determine these weights for the \textit{CPI} method applied in this evaluation, a grid search is conducted.
More specifically, several combinations of the three weights $w = (w_e, w_w, w_s)$ are tested with $w_e$ in range $[1 \twodots 20]$, $w_w$ in range $[0 \twodots 10]$, and $w_s$ in range $[1 \twodots 20]$ on a calibration set.
This calibration set consists of five time series with characteristics similar to the 50 time series used for the evaluation.
Based on the grid search, the weights $w = (5, 1, 10)$ are selected for the \textit{CPI} method as they provide good and robust results regarding the $\mathrm{MAPE_p}$~(\ref{eq:MAPE_p}).

\subsection{Results}
\label{subsec:results}

Based on the 50 selected time series, the selected benchmark methods, and the experimental setup described above, this section explains the results.
It first covers the use of matching patterns and the conservation of energy, quantified by $\mathrm{MAPE_p}$~(\ref{eq:MAPE_p}) and $\mathrm{WAPE_E}$~(\ref{eq:WAPE_E}) respectively.
The presented values of these error measures are the truncated means for the 50 evaluated time series, which omit the two best and worst values to obtain less outlier-sensitive results.
Afterward, the run-time of the evaluated methods is addressed.
Lastly, the output of an exemplary imputation visually illustrates the evaluation results.

\subsubsection{Use of Matching Patterns}
The use of matching patterns -- as defined in \autoref{eq:MAPE_p} -- by the evaluated methods is presented in \autoref{fig:mapep}. 
For the six different shares of artificially inserted missing values, the figure shows the $\mathrm{MAPE_p}$~(\ref{eq:MAPE_p}) of all evaluated methods.
Regardless of the share of missing values, the \textit{CPI} method performs best.
Compared to the historical average -- the best benchmark method --, the average $\mathrm{MAPE_p}$~(\ref{eq:MAPE_p}) of the \textit{CPI} method is 8.27\% better. 
The linear interpolation performs worst for all shares of missing values.
The historical average and the \textit{Prophet}-based method perform considerably better than the linear interpolation, with the historical average performing slightly better for all shares.
All methods tend to higher errors with higher shares of missing values. 
This trend is weakest for the historical average and most pronounced for the linear interpolation. 
With regard of the errors of individual time series, the benchmark methods are more prone to extreme errors with a maximum $\mathrm{MAPE_p}$~(\ref{eq:MAPE_p}) of 9.75 and higher while the \textit{CPI} method has a maximum $\mathrm{MAPE_p}$~(\ref{eq:MAPE_p}) of 2.28.

\subsubsection{Conservation of Energy}
The conservation of energy for each gap -- as defined in \autoref{eq:WAPE_E} -- is shown in \autoref{fig:mapee} for the four evaluated methods.
The figure presents the $\mathrm{WAPE_E}$~(\ref{eq:WAPE_E}) for the six shares of missing values.
Again, the \textit{CPI} method performs best regardless of the share of missing values.
To allow a better comparability with the benchmark methods that all do not use scaling, the dashed line indicates the error values for the \textit{CPI} method without scaling.
Without scaling, the \textit{CPI} method performs on average 4.9\% better than the second best method.
The scaling further reduces this error to nearly zero for all shares of missing values such that the  \textit{CPI} method performs even better.
The linear interpolation again performs worst for all shares of missing values.
However, the \textit{Prophet}-based method performs better than the historical average for this metric.
Similar to the previous error measure, the benchmark methods yield more extreme error values with a maximum $\mathrm{WAPE_E}$~(\ref{eq:WAPE_E}) of at least 2.47 while the maximum of the \textit{CPI} method is 0.84 without scaling.

\begin{figure}
	\begin{center}
		\ref*{named1}		
	\end{center}
	\begin{tikzpicture}
		\begin{axis}[
			name=main plot,
			width=0.99\linewidth,
			height=5.5cm,
			xlabel={Share of missing values},
			ylabel={$\mathrm{MAPE_p}$},
			enlarge x limits=0.05,
			enlarge y limits=0.1,
			xtick={0.01,0.05,0.1,0.2,0.3},
			xticklabel={\pgfmathparse{\tick*100}\pgfmathprintnumber{\pgfmathresult}\%},
			legend style={
				legend cell align=center,
				legend columns=4,
				legend to name=named1,
				draw=none,
			},
		]
		
			\addplot[color4,mark=triangle*] table[col sep=comma,x index=0,y index=6] {data/evaluation/trimmedmean/cpi-1-10-5plus.csv};
			\addplot[color3,mark=halfsquare*] table[col sep=comma,x index=0,y index=6] {data/evaluation/trimmedmean/prophet.csv};
			\addplot[color2,mark=*] table[col sep=comma,x index=0,y index=6] {data/evaluation/trimmedmean/histavg.csv};
			\addplot[color5,mark=pentagon*] table[col sep=comma,x index=0,y index=6] {data/evaluation/trimmedmean/linear.csv};
			\legend{CPI,Prophet,HistAvg,Linear}
		\end{axis}
	\end{tikzpicture}
	\caption{
		The $\mathrm{MAPE_p}$~(\ref{eq:MAPE_p}) of the \textit{CPI} method and the three benchmark methods with different shares of missing values.
		As the scaling of imputed values does not noticeably affect the results of the \textit{CPI} method, it is omitted in this figure.
	}
	\label{fig:mapep}

	\vskip 12pt

	\begin{center}
		\ref*{named2}		
	\end{center}
	\begin{tikzpicture}
		\begin{axis}[
			anchor=north west,
			width=0.99\linewidth,
			height=5.5cm,
			xlabel={Share of missing values},
			ylabel={$\mathrm{WAPE_E}$},
			yticklabel style={/pgf/number format/.cd,fixed,precision=2},
			enlarge x limits=0.05,
			enlarge y limits=0.1,
			xtick={0.01,0.05,0.1,0.2,0.3},
			xticklabel={\pgfmathparse{\tick*100}\pgfmathprintnumber{\pgfmathresult}\%},
			legend style={
				legend cell align=center,
				legend columns=1,
				legend to name=named2,
				draw=none,
			},
		]
			\addplot[color4,mark=triangle*,dashed] table[col sep=comma,x index=0,y index=1] {data/evaluation/trimmedmean/cpi-1-10-5.csv};
			\addplot[color4,mark=triangle*] table[col sep=comma,x index=0,y index=1] {data/evaluation/trimmedmean/cpi-1-10-5plus.csv};
			\addplot[color3,mark=halfsquare*] table[col sep=comma,x index=0,y index=1] {data/evaluation/trimmedmean/prophet.csv};
			\addplot[color2,mark=*] table[col sep=comma,x index=0,y index=1] {data/evaluation/trimmedmean/histavg.csv};
			\addplot[color5,mark=pentagon*] table[col sep=comma,x index=0,y index=1] {data/evaluation/trimmedmean/linear.csv};
			\legend{CPI without scaling}
		\end{axis}
	\end{tikzpicture}
	\caption{
		The $\mathrm{WAPE_E}$~(\ref{eq:WAPE_E}) of the CPI method and the three benchmark methods with different shares of missing values.
		For better comparability with the benchmark methods that all do not use scaling, the dashed line indicates the $\mathrm{WAPE_E}$ of the \textit{CPI} method without scaling the imputed values to preserve the energy of a gap.
	}
	\label{fig:mapee}
	
	\vskip 16pt
		
	\begin{tikzpicture}
		\begin{semilogyaxis}[
			anchor=north west,
			width=0.99\linewidth,
			height=5.5cm,
			xlabel={Share of missing values},
			ylabel={Run-time in seconds},
			yticklabel style={/pgf/number format/.cd,fixed,precision=2},
			enlarge x limits=0.05,
			enlarge y limits=0.1,
			xtick={0.01,0.05,0.1,0.2,0.3},
			xticklabel={\pgfmathparse{\tick*100}\pgfmathprintnumber{\pgfmathresult}\%},
			]
			\addplot[color4,mark=triangle*] table[col sep=comma,x index=0,y index=7] {data/evaluation/trimmedmean/cpi-1-10-5plus.csv};
			\addplot[color3,mark=halfsquare*] table[col sep=comma,x index=0,y index=7] {data/evaluation/trimmedmean/prophet.csv};
			\addplot[color2,mark=*] table[col sep=comma,x index=0,y index=7] {data/evaluation/trimmedmean/histavg.csv};
			\addplot[color5,mark=pentagon*] table[col sep=comma,x index=0,y index=7] {data/evaluation/trimmedmean/linear.csv};
		\end{semilogyaxis}
	\end{tikzpicture}
	\caption{
		The average run-times required by the \textit{CPI} method and the three benchmark methods for the imputation of the 50 selected one-year time series.
		Note the logarithmic time scale, which visually compresses \textit{Prophet}'s run-time decrease by 7 seconds from 1 to 30\% of missing values.
	}
	\label{fig:timeonly}
\end{figure}

\subsubsection{Run-time}
\autoref{fig:timeonly} shows the average run-times required by the evaluated methods for the imputation of the 50 selected one-year time series with 35,040 values each.
The linear interpolation and the historical average methods are by far the fastest methods.
The \textit{CPI} method requires about 10 to 20 times more time than these two methods.
The \textit{Prophet}-based method requires much more time than the other methods and is more than 10 times slower than the introduced \textit{CPI} method.

\begin{figure}
	\begin{tikzpicture}
		\begin{semilogxaxis}[
			anchor=north west,
			width=0.99\linewidth,
			height=5.5cm,
			xlabel={Run-time in seconds},
			ylabel={$\mathrm{MAPE_p}$},
			yticklabel style={/pgf/number format/.cd,fixed,precision=2},
			enlarge x limits=0.05,
			enlarge y limits=0.1,
			]
			\addplot[color4,mark=triangle*,only marks] table[col sep=comma,x index=7,y index=6] {data/evaluation/trimmedmean/cpi-1-10-5plus.csv};
			\addplot[color3,mark=halfsquare*,only marks] table[col sep=comma,x index=7,y index=6] {data/evaluation/trimmedmean/prophet.csv};
			\addplot[color2,mark=*,only marks] table[col sep=comma,x index=7,y index=6] {data/evaluation/trimmedmean/histavg.csv};
			\addplot[color5,mark=pentagon*,only marks] table[col sep=comma,x index=7,y index=6] {data/evaluation/trimmedmean/linear.csv};
		\end{semilogxaxis}
		
	\end{tikzpicture}	
	\caption{
		Comparison of the use of matching patterns and the average run-time needed of the \textit{CPI} method and the three benchmark methods for the imputation of 50 one-year time series.
		The x-axis shows the required average run-times and the y-axis the $\mathrm{MAPE_p}$~(\ref{eq:MAPE_p}).
	}
	\label{fig:time}
\end{figure}
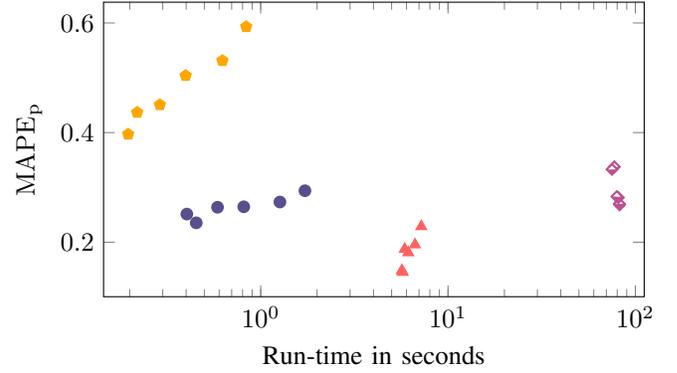

\subsubsection{Use of Matching Patterns vs. Run-time}
In \autoref{fig:time}, the obtained results regarding the use of matching patterns is put in relation to the run-time needed with a scatter plot showing the required average run-times on the x-axis and the $\mathrm{MAPE_p}$~(\ref{eq:MAPE_p}) on the y-axis.
Smaller values indicate a better performance.
While the linear interpolation and historical average method provide fast and inaccurate results, the \textit{CPI} method delivers the most accurate results with a reasonable run-time.
The \textit{Prophet}-based method yields mediocre results while taking much longer to calculate than the other methods.

\subsubsection{Exemplary Imputation Results}
\begin{figure*}
	\centering	
	\includegraphics[width=\linewidth]{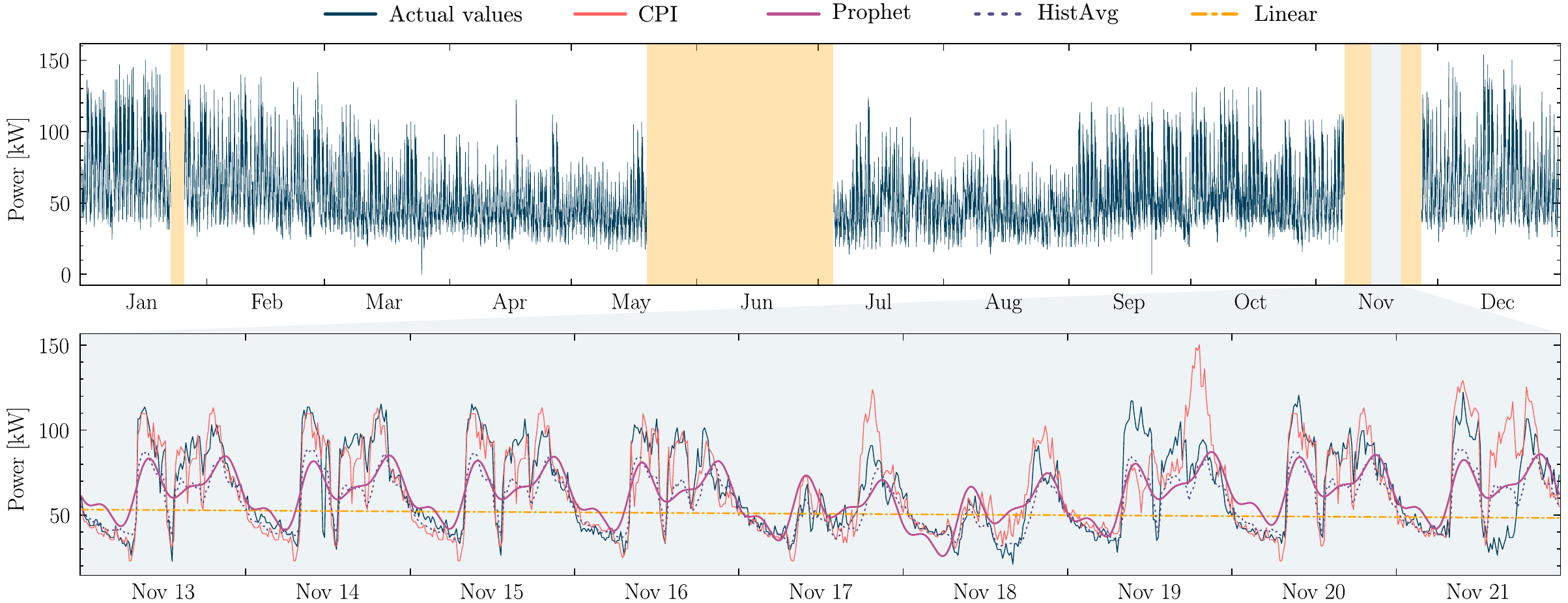}
	\caption{
		The upper figure shows an exemplary one-year time series with 20\% of missing values.
		For a multi-day excerpt of a gap in November, the lower figure presents the resulting imputations by the \textit{CPI} method and the three benchmark methods in comparison to the actual values.
		Metrics of the \textit{CPI} method and the best benchmark method for this example:
		$\mathrm{MAPE_p}$~(\ref{eq:MAPE_p}): 0.188 (Linear: 0.271), $\mathrm{WAPE_E}$~(\ref{eq:WAPE_E}): 0.003 (\textit{Prophet}: 0.073), run-time: 6.75s (Linear: 0.64s)
	}
	\label{fig:cpi-result}
\end{figure*}

For all evaluated imputation methods, \autoref{fig:cpi-result} visually illustrates an exemplary imputation of a time series with 20\% of artificially inserted missing values, resulting in large gaps.
The imputation of the linear interpolation fails to capture the patterns of the time series.
The imputations by the historical average method and the \textit{Prophet}-based method capture the essential patterns but lack details.
The imputation by the \textit{CPI} method mostly fits the actual values but it shifts and increases some peaks.
Compared to the three benchmark methods, the imputation by the novel \textit{CPI} method comes closest to the actual values.

\section{Conclusion and Outlook}
\label{sec:conclusion}

The present paper introduces a new \textit{Copy-Paste Imputation} method for energy time series.
It copies blocks of data with similar properties and pastes them into gaps of the time series.
This approach enables realistic imputations even for large gaps with several weeks of consecutively missing values.
In contrast to all other methods in the literature -- to the best knowledge of the authors --, the \textit{CPI} method utilizes the often provided \textit{energy} time series, i.e.~the actual meter readings, instead of \textit{power} time series, i.e.~the average power per interval.
Using energy time series allows for a robust selection of matching blocks of data and ensures that the overall energy per gap remains unchanged while imputing the missing values with realistic patterns.
For the imputation, the \textit{CPI} method does not require additional information such as weather data or energy/power time series of related smart meters.

The proposed \textit{CPI} method is applied to a real-world dataset and compared to three benchmark methods.
For the evaluation, six shares of artificially inserted missing values between 1 and 30\% are used.
For all shares of missing values, the \textit{CPI} method clearly outperforms the benchmark methods.
The evaluation confirms that the \textit{CPI} method uses matching patterns for the imputations and that it conserves the overall energy of every imputed gap while requiring only a moderate run-time.

Based on these results, \textit{future work} could follow three directions. 
First, the robustness could be analyzed and improved regarding aperiodic events or time series with other periodicities or temporal resolutions.
Second, a trend analysis could enhance the selection of matching days especially for longer gaps.
Third, anomaly or error detection functions could be integrated to repair implausible values.
Moreover, a reporting and analysis tool could use the \textit{CPI} method to estimate the imputation quality based on artificially inserted missing values.

\bibliography{min}

\end{document}